\documentclass[runningheads]{llncs}

\usepackage{eccv}

\usepackage{eccvabbrv}

\usepackage{graphicx}
\graphicspath{{./}{VidParse_ECCV2026/}}  %
\usepackage{booktabs}
\usepackage{colortbl}
\usepackage{bbding}
\usepackage[dvipsnames,table,xcdraw]{xcolor}

\usepackage[accsupp]{axessibility}  %
\usepackage[dvipsnames]{xcolor} %

\newcommand{\method}{VidParse\xspace}

\usepackage{hyperref}
\usepackage{adjustbox}

\usepackage{orcidlink}

\begin{document}

\setcounter{tocdepth}{2}          %
\addtocontents{toc}{\protect\setcounter{tocdepth}{-5}}  %

\title{VidParse: Online Parsing of Egocentric Procedures Like a Pro} 

\titlerunning{VidParse}

\author{Anubhav Gupta \and
Archit Kambhamettu \and
Vatsal Agarwal \and
Pulkit Kumar \and
Abhinav Shrivastava}

\authorrunning{A.~Gupta et al.}

\institute{University of Maryland, College Park, USA\\
\email{\{anubhav, architk, vatsalag, pulkit, abhinav2\}@umd.edu}}
\pdfbookmark[0]{VidParse: Online Parsing of Egocentric Procedures Like a Pro}{maintitle}
{
\renewcommand{\addcontentsline}[3]{}
\maketitle
}

\begin{abstract}

  Translating continuous, noisy egocentric video streams into discrete, temporally ordered action steps is fraught with visual challenges. Heavy ego-motion, transient occlusions, and the high intra-class variability of unscripted human-object interactions cause standard frame-level online temporal models to struggle, often resulting in severe over-segmentation and structural collapse. To bridge the gap between unstable low-level perception and high-level procedural logic, we present \emph{VidParse}, a online, training-free framework that treats activity understanding as a graph-constrained inference problem. Rather than relying on learned temporal filters, we dynamically identify semantic transitions using a temporal similarity matrix over manipulation-anchored features, which are extracted from frozen foundation models to prioritize foreground hand-object interactions. A beam search decoder then leverages an induced procedural task graph to explicitly enforce valid action transitions and prune impossible trajectories. By anchoring robust visual segments to hard procedural constraints, our approach preserves long-range state transitions and achieves up to a 10x improvement in complex multi-step parsing accuracy over strong online baselines, all without requiring a single gradient update.
  
  \noindent\textbf{Project Page \& Code:} \url{https://learn2phoenix.github.io/VidParse}
\end{abstract}

\begin{figure}[t]
    \centering
    \includegraphics[width=0.9\linewidth]{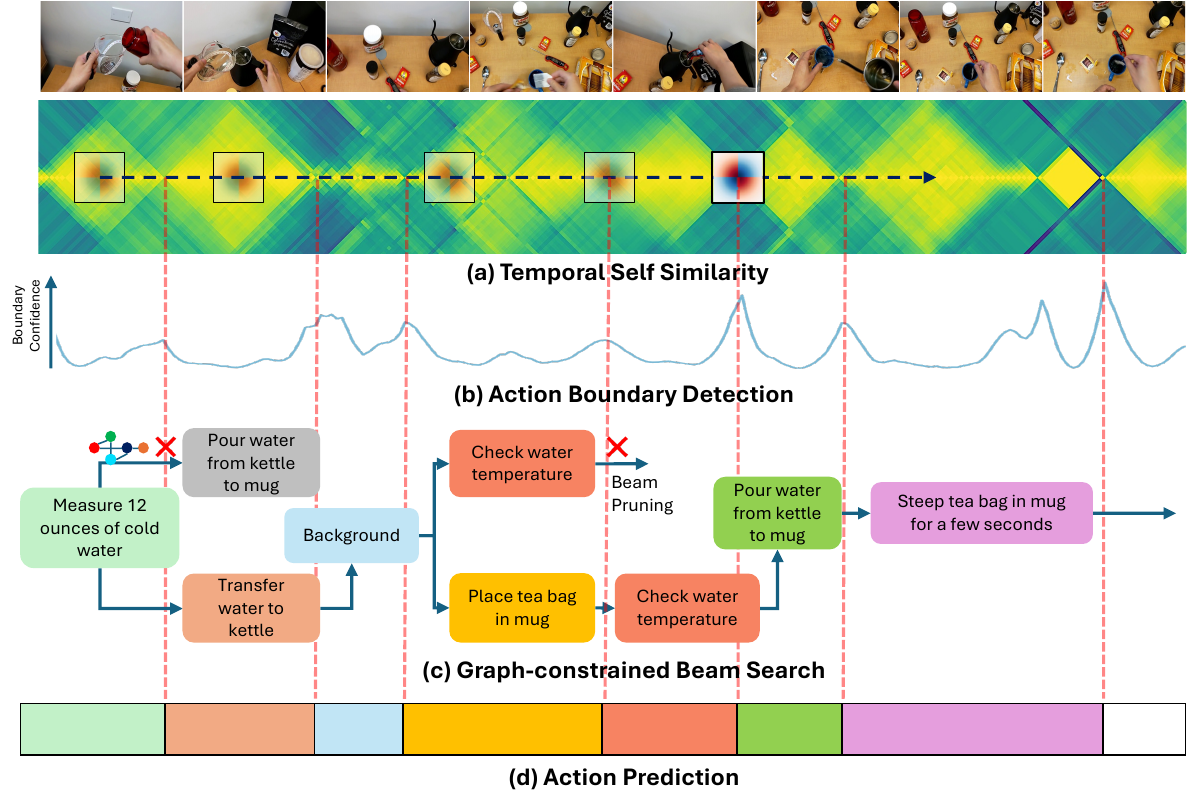}
    \caption{\textbf{VidParse:} (a) \textit{Object-Centric Self-Similarity}: We compute a Temporal Similarity Matrix (TSM) strip from frozen DINOv2 features. A Gaussian-tapered checkerboard kernel (red/blue overlay) slides along the diagonal to detect semantic transitions (b) \textit{Training-Free Boundary Detection}: peaks indicate likely action boundaries. (c) \textit{Graph-Constrained Beam Search}: These boundaries trigger a beam search that assigns action labels (d) \textit{Final Prediction}: The result is a coherent, causally valid segmentation.}
    \label{fig:teaser}
    
\end{figure}

\section{Introduction}
\label{sec:intro}

Egocentric procedure parsing aims to convert a streaming first-person video into an ordered sequence of action steps as a person performs a task. In many real-world applications, this reasoning must occur online as the video unfolds. Unlike offline temporal segmentation, which can rely on full video context to resolve ambiguities, the online setting must infer the current step using only past observations. Maintaining a coherent estimate of procedural state over long sequences is therefore the central challenge.

Most existing approaches address this problem by learning temporal models that capture action transitions from annotated training data. Progress-aware methods explicitly model task completion through structured state representations, while sequence models learn transition dynamics between actions using large amounts of labeled data. Although effective in controlled settings, these approaches depend heavily on training data and learned temporal dynamics, making them sensitive to domain shifts and difficult to deploy in settings where annotated data or retraining is impractical.

Rather than learning temporal models, we formulate egocentric activity understanding as a structured inference problem governed by procedural constraints. Many everyday tasks follow consistent structural dependencies: certain actions must precede others, while others cannot occur simultaneously. Exploiting these dependencies allows action sequences to be inferred even when local visual evidence is ambiguous.

Inspired by this, we introduce \textbf{VidParse}, a \textit{training-free structured parsing framework} for online egocentric procedure understanding. We define ``training-free'' to indicate that our method requires no task-specific training or fine-tuning on the downstream procedure parsing dataset; instead, it relies entirely on frozen pretrained components (e.g., DINOv2 and hand-object detectors). Our approach combines three complementary components as shown in Fig.~\ref{fig:teaser}. First, we construct interaction-focused visual representations using Manipulation-Anchored Features (MAFs), which combine frozen DINO descriptors with hand-object detections to emphasize regions where interactions occur. Second, we detect action boundaries online by computing a temporal similarity matrix over these features and applying a classical checkerboard kernel, enabling segmentation without learned temporal filters or boundary predictors. Finally, we perform structured decoding over an induced procedural task graph using graph-constrained beam search, selecting the action sequence that best satisfies both visual evidence and procedural constraints while maintaining causal inference.

We evaluate our approach on two challenging egocentric procedure parsing benchmarks, GTEA and EgoPER, which exhibit substantially different task structures and execution variability. 
Despite requiring no training or fine-tuning, our framework performs competitively with strong online baselines while achieving improved temporal consistency. 

Beyond standard segmentation metrics, we introduce an $N$-step transition evaluation that measures long-range procedural consistency by analyzing multi-step action dependencies, enabling evaluation of structured procedural predictions.
Our method maintains coherent multi-step trajectories more reliably than prior online approaches, demonstrating stronger preservation of global procedural structure. In summary, our contributions are as follows:
\begin{itemize}
\item \textbf{Training-free online procedure parsing.}
We introduce \textbf{VidParse}, a framework for causal egocentric procedure parsing that operates without learning temporal models or fine-tuning foundation networks.

\item \textbf{Structured procedural inference.}
We formulate online procedure parsing as structured inference over an induced task graph and perform graph-constrained beam search to enforce valid action transitions.

\item \textbf{Long-range procedural evaluation.}
We introduce an $N$-step transition metric that measures the consistency of multi-step action dependencies beyond standard segmentation metrics.

\item \textbf{Manipulation-anchored action segmentation.}
We introduce Manipulation-Anchored Features (MAFs) and detect semantic action boundaries using a checkerboard kernel applied to an online temporal similarity matrix.

\end{itemize}

\section{Related Work}
Our work sits at the intersection of online temporal action segmentation, structure-aware (grammar/graph) procedural parsing, and similarity-based boundary detection. We briefly review each line of work and highlight how our method combines frozen foundation model features with non-parametric inference.

\subsection{Temporal Action Segmentation}

\begin{figure*}[t]
    \centering
    \includegraphics[width=\textwidth]{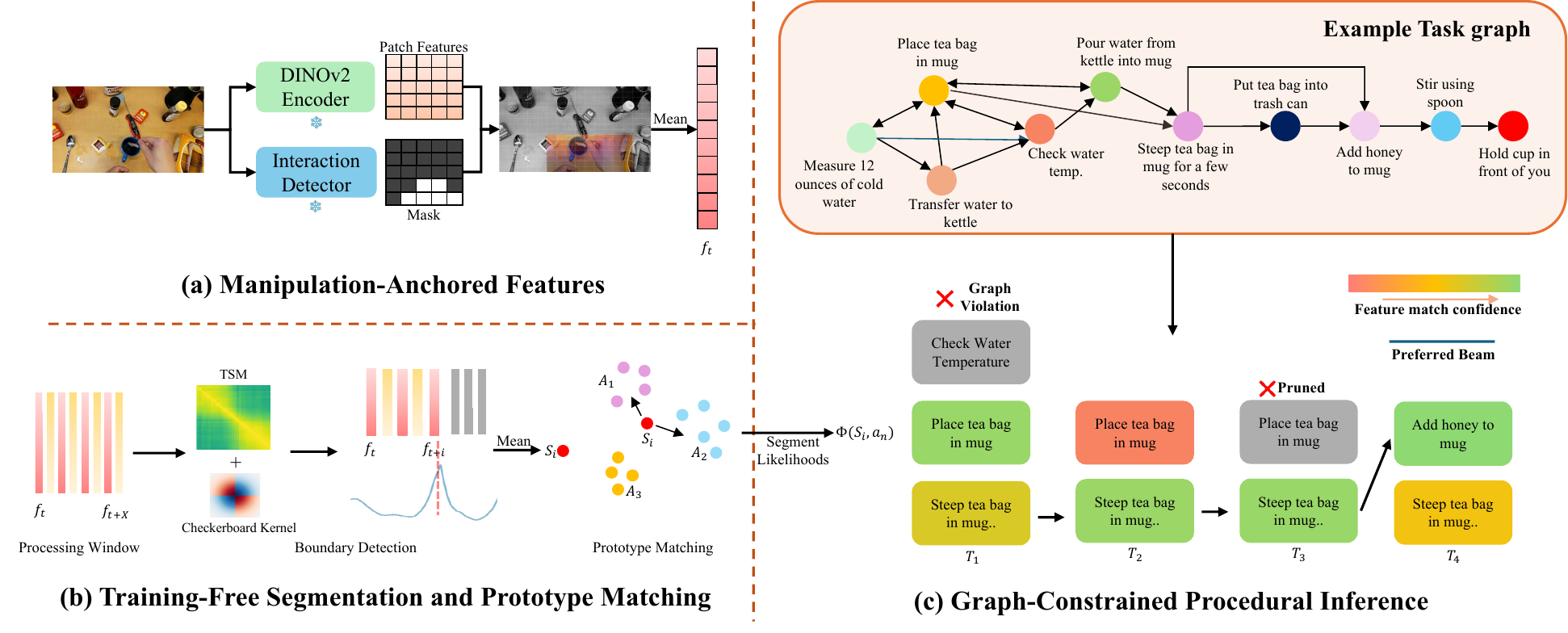}
    \caption{\textbf{VidParse Framework:} (a) \textit{Manipulation-Anchored Features}: We compute spatial features from a frozen DINOv2 backbone and apply a binary mask derived from a Hand-Object Detector (HOD)(b) \textit{Training-Free Segmentation and Prototype Matching}: A sliding-window Temporal Similarity Matrix (TSM) is convolved with a Gaussian-tapered checkerboard kernel to detect semantic boundaries online. The resulting segments ($S_i$) are scored against pre-computed Action Centroids (c) \textit{Graph-constrained Procedural Inference}: We infer the optimal sequence using a beam search that strictly enforces transitions defined in the induced Task Graph. Transitions that violate the procedural structure (e.g., pruned path at $T_1$) are assigned infinite cost and removed, ensuring causally valid segmentation.}
    \label{fig:mainfig}
    
\end{figure*}

Temporal Action Segmentation (TAS) has evolved from offline models that utilize full video context (e.g., MS-TCN \cite{farha2019ms}, ASFormer \cite{yi2021asformer}) to causal Online Action Segmentation (OAS) \cite{zhong2024onlinetas} required for real-time egocentric applications. While the field initially focused on third-person videos, with datasets like 50 Salads \cite{stein2013combining}, Breakfast \cite{kuehne2014language}, Kinetics \cite{kay2017kinetics}, Something-Something \cite{goyal2017something}, Cross-Task \cite{zhukov2019cross}, COIN \cite{tang2019coin}, recent focus has shifted to the egocentric perspective. With datasets like EPIC-Kitchens \cite{damen2020epic}, Ego4D \cite{grauman2022ego4d}, CaptainCook4D \cite{peddi2024captaincook4d}, EgoPER \cite{lee2024error}, and HD-EPIC \cite{perrett2025hd}, the community is now addressing extreme ego-motion, occlusion, and long-tail distributions. Egocentric tasks like cooking and assembly often require real-time reasoning, necessitating Online Action Segmentation (OAS), where models must reason causally using only past frames. Without future context, online frame-level classifiers often suffer from severe over-segmentation. Recent mitigations include adaptive memory banks (OnlineTAS \cite{zhong2024onlinetas}, MATR \cite{song2024online}, sub-quadratic state-space models (Mamba-OTR \cite{catinello2025mamba}), and test-time vision-language adaptations (T3AL \cite{liberatori2024test}, FreeZAD \cite{han2025training}). While Large Multimodal Models offer broad generalization, their computational cost prohibits real-time dense segmentation. Our work bridges this gap by extracting semantic robustness directly from frozen foundation models within a lightweight, training-free framework, bypassing complex temporal optimization entirely.

\subsection{Similarity-Based Boundary Detection} 
Generic Event Boundary Detection (GEBD) identifies semantic transitions without predefined taxonomies. Recent advancements replace classical techniques with predictive cognitive models (ESTimator \cite{jung2025online}), dynamic multi-exit architectures (DyBDet \cite{zheng2024fine}), and denoising diffusion processes (DiffGEBD \cite{hwang2025diffgebd}). Despite eliminating dense frame-level labels, unsupervised (UBoCo \cite{kang2022uboco}, OTAS \cite{li2024otas}) and weakly-supervised (ATBA \cite{xu2024efficient}) methods still require substantial feature or decoder optimization. We demonstrate that the inherent semantic stability of frozen foundation models like DINOv2 \cite{oquab2023dinov2} revitalizes classical signal processing. By applying a Gaussian-tapered checkerboard kernel \cite{foote2000automatic, cooper2001scene} directly to object-centric patch embeddings, we achieve high-fidelity boundary emission. To our knowledge, this is the first to seamlessly integrate training-free GEBD into a graph-constrained action parsing pipeline.

\subsection{Grammar-Based and Structural Parsing}
To handle minutes-long procedural dependencies, recent works increasingly inject structural priors via neuro-symbolic reasoning. A foundational work in this domain is VideoGraph \cite{hussein2019videograph}, which was among the first to model minutes-long human activities using a learned graph structure. Unlike purely sequential models, VideoGraph constructs a soft, undirected graph of latent concepts (unit-actions), demonstrating that modeling the structural relationships between atomic actions is essential for understanding long-form videos. Building on this structural intuition, subsequent works model activities using induced grammars~\cite{gong2023activity}, learn soft task graphs for progress estimation \cite{shen2024progress}, or leverage large language models to extract logic for spatio-temporal alignment (LASER \cite{huang2023laser}, PHGC \cite{jiang2025phgc}, KML \cite{nguyen2026pkr}).

Similarly, GTG2Vid \cite{lee2025error} aligns videos with generalized task graphs for error detection. Other recent works explore temporal correspondence to segment procedural steps (e.g., \cite{bansal2022my, chowdhury2024opel, peirone2025hiero, ali2025joint}); however, these are predominantly offline procedure learning and alignment methods. Our setting differs fundamentally as we focus on online, causal parsing from streaming video. Furthermore, while most prior works employ graphs as soft regularizers during training, we integrate the induced task graph as a strict, hard constraint during beam search. By assigning infinite costs to invalid procedural transitions, our decoder explicitly prevents the structural collapse endemic to local predictors.

\noindent\textbf{Summary and Positioning:} Prior work typically focuses on (i) online temporal segmentation with learned temporal models, (ii) structural priors via soft graph or grammar regularization, or (iii) boundary detection through trained or optimized objectives. Our framework combines these threads by using a classical, training-free boundary detector to emit action segments, and then enforcing procedure as a hard constraint during inference.

\section{Method}
\label{sec:method}

We propose \method, a training-free framework for online egocentric procedure parsing. Given a first-person video, the goal is to convert it into a sequence of discrete action steps while respecting the procedural structure of the task.

\method proceeds in four stages. 
First, we extract interaction-focused visual representations called Manipulation-Anchored Features (MAFs) that isolate hand-object interactions using frozen foundation models. 
Second, we detect action boundaries online using a training-free temporal similarity detector, which identifies transitions between manipulation phases. 
Third, we assign action labels to segments through prototype-based matching that maps visual segments to semantic actions without training. Finally, we decode the resulting sequence through structured procedural inference using graph-constrained beam search, which enforces valid task transitions.

\subsection{Problem Formulation}
\label{subsec:problem_formulation}

Given a streaming egocentric video, the objective is to infer an ordered sequence of procedural actions online while respecting task constraints. We formulate online procedure parsing as a Maximum A Posteriori (MAP) inference problem over a sequence of temporal segments.

Formally, let $\mathcal{V}$ denote a streaming video. At each time step $t$, we extract a frame-level visual representation $\mathbf{v}_t \in \mathbb{R}^d$. An online boundary detector $\phi_{boundary}$ operates over these features and partitions the stream into $N$ segments
\[
\mathcal{S} = \{s_1, s_2, \dots, s_N\}.
\]

Each segment $s_i$ corresponds to a contiguous block of frame-level features representing a single procedural step.

Our objective is to assign an action label to each segment while respecting the structure encoded by a procedural graph $\mathcal{G}$. The optimal sequence of actions
is obtained by solving
\begin{equation}
\mathbf{A}^* = \arg \max_{\mathbf{A}} P(\mathbf{A} \mid \mathcal{S}, \mathcal{G}).
\end{equation}

\subsection{Manipulation-Anchored Features (MAFs) }
\label{subsec:maf}

Actions in egocentric tasks are primarily defined by how the hands interact with objects.  Consequently, the most informative visual signal lies in regions where these interactions occur rather than in the entire scene.
This echoes a broader finding in video representation learning, where confining tokens to task-relevant trajectories rather than the full frame yields representations that isolate motion from scene appearance~\cite{kumar2024trajectory,kumar2025trokens}; we apply the same principle spatially, anchoring the representation to the hand--object envelope. To capture this signal, we construct Manipulation-Anchored Features (MAFs), which focus the representation on these interaction regions while suppressing background clutter.

We begin by extracting patch-level descriptors using a frozen DINOv2 backbone \cite{oquab2023dinov2}. Specifically, we use the patch tokens from the final layer of a ViT-L/14 encoder \cite{dosovitskiy2020image}.
To localize the interaction region, we use a frozen Hand-Object Detector (HOD) \cite{shan2020understanding} to detect hands and manipulated objects in each frame. The detected bounding boxes are merged using an enclosing strategy (ref.~\ref{subsec:impl_dets}) that produces a single bounding region covering the interaction area.

This region is projected onto the patch grid to define a spatial slice
\[R_t = [p_{x1}, p_{y1}, p_{x2}, p_{y2}]\]

The frame-level representation is then computed as the spatial average of patch tokens within this region:
\begin{equation}
\mathbf{f}_t =
\frac{1}{|R_t|}
\sum_{i=p_{y1}}^{p_{y2}}
\sum_{j=p_{x1}}^{p_{x2}}
\mathbf{P}_{t,i,j}.
\end{equation}

If no detections are present in a frame, we propagate the last valid feature forward. This simple temporal smoothing prevents head motion or short occlusions from introducing spurious changes in the feature stream.

The resulting interaction-focused descriptors form the \textbf{Manipulation Anchored Features (MAFs)} used throughout the remainder of the pipeline.

\subsection{Training-Free Boundary Detection}
\label{subsec:obd}

Procedural parsing requires identifying transitions between consecutive actions. 
Rather than learning a boundary predictor, we detect transitions directly from the temporal similarity structure of the MAF feature stream using a training-free novelty detector.

Our approach follows three steps: constructing a local similarity matrix over a sliding temporal window, computing a novelty signal using a checkerboard kernel, and emitting boundaries through online peak detection. This ensures that the system remains ``online,'' operating only on a short buffer of the most recent frames rather than the entire video sequence.

\noindent\textbf{Gaussian-tapered checkerboard kernel.} The use of a Gaussian-tapered checkerboard kernel to detect boundaries in a Self-Similarity Matrix (SSM) was pioneered in \cite{foote2000automatic} and \cite{cooper2001scene} where they demonstrated that transitions between homogeneous signal segments manifest as checkerboard patterns along the SSM diagonal. The core of our detector is a checkerboard kernel $\mathbf{K} \in \mathbb{R}^{2L \times 2L}$, where $L$ defines the temporal horizon of the sliding window. To emphasize frames near the center of the window and suppress noise at the edges, we apply a radial Gaussian taper to the kernel. The kernel is defined as
\[
\mathbf{K} = (\mathbf{C} \otimes \mathbf{J}_L) \odot \mathbf{G},
\]

where
\[
\mathbf{C} =
\begin{bmatrix}
1 & -1 \\
-1 & 1
\end{bmatrix}
\]

\noindent is the base checkerboard pattern, $\mathbf{J}_L$ is an $L \times L$ matrix of ones, and $\mathbf{G}$ is a Gaussian weighting matrix centered at the kernel midpoint.

\noindent\textbf{Local novelty computation.} At each time step we maintain a buffer containing the most recent $2L$ MAF features. Using this buffer we compute a local self-similarity matrix $\mathbf{S} \in \mathbb{R}^{2L \times 2L}$, whose entries measure cosine similarity between normalized features:
\begin{equation}
S_{i,j} =
\frac{\mathbf{f}_i^\top \mathbf{f}_j}
{\|\mathbf{f}_i\| \|\mathbf{f}_j\|}.
\end{equation}

\noindent The novelty score at time $t$ is obtained through the Frobenius inner product of this matrix with the kernel
\begin{equation}
\Delta_t =
\sum_{i=1}^{2L}
\sum_{j=1}^{2L}
S_{i,j} K_{i,j}.
\end{equation}

Large novelty values indicate abrupt changes in the  structure and therefore suggest an action transition. Because DINOv2 is trained to be semantically stable, an action (e.g., ``chopping'') appears as a homogeneous block in the MAF, allowing linear filters to perform complex semantic segmentation tasks that previously required deep temporal networks.

\noindent\textbf{Boundary emission.} The novelty signal is converted into discrete boundaries using online peak detection. A boundary is emitted when the novelty score forms a local maximum and exceeds a saliency threshold $h$:

\begin{equation}
\tau =
\{t \mid \Delta_t > \Delta_{t \pm k} \text{ and } \Delta_t > h\}.
\end{equation}

To avoid over-segmentation, we enforce a minimum temporal distance $d$ from the previous boundary. Once a peak is detected, the buffer is flushed up to index $\tau$, producing a segment that is passed to the inference stage.

\subsection{Prototype-Based Action Matching}
\label{subsec:matching}

To assign semantic labels to the detected segments, we employ a non-parametric, micro-prototype matching scheme. A na\"ive approach would average all instances of an action into a single global prototype. However, procedural actions are often lengthy and multi-phasic; averaging an entire execution sequence blurs distinct sub-states and dilutes the Manipulation-Anchored Feature (MAF) signal. To preserve local temporal semantics, we first group training examples into execution styles using agglomerative clustering and compute representative centroids. We then temporally slice each clustered action sequence representations into overlapping, short-duration windows (e.g., 2 seconds). This produces a dense library of \textit{micro-prototypes}, denoted as $\mathcal{P}_a$, capturing the distinct temporal phases of each action $a$ without requiring temporal warping. During online inference, when the boundary detector emits a segment $s_k$, we apply the Hand-Object Detector (HOD) mask to remove frames without valid interactions. We compute the mean descriptor $\mathbf{g}_k$ over this filtered segment and assign labels by finding the nearest neighbor in the micro-prototype space. The matching cost between the segment and action $a$ is defined by the minimum cosine distance to its micro-prototypes:
\begin{equation}
\phi(s_k, a) =
\min_{p \in \mathcal{P}_a}
\left(
1 - \frac{\mathbf{g}_k^\top p}
{\|\mathbf{g}_k\| \|p\|}
\right).
\end{equation}

\subsection{Structured Procedural Inference}
\label{subsec:graph}

Visual evidence alone can lead to ambiguous or noisy action assignments, particularly when multiple actions exhibit similar visual patterns. 
To enforce valid task progression, we perform structured procedural inference using a task graph that encodes allowable action transitions. 
We formulate decoding as the minimization of a structured energy function that integrates visual evidence, manipulation cues, and graph constraints. 
Finally, we incorporate a rectification mechanism that handles short visual gaps caused by occlusions or detector failures.

\noindent\textbf{Task graph construction.} We induce the task-graph from training sequences by aggregating step-ordered annotations and identifying prerequisite relationships between actions. First, we record all immediate sequential transitions, distinguishing between first-time visits and revisits. 
We then accumulate previously observed steps as potential prerequisites for each action’s first occurrence. We also designate steps as omittable if valid instances exist without that step in the training set. Finally, actions with an empty minimal prerequisite set are classified as start nodes, and we fully connect all start nodes to each other

\noindent\textbf{Energy formulation.} We combine visual evidence, manipulation cues, and task-graph constraints to score candidate labelings.

\begin{equation}
\label{eq:energy}
\mathbf{A}^* =
\arg \min_{\mathbf{A}}
\sum_{i=1}^{N}
\Big[
D_i \cdot \phi(s_i, a_i)
+ \lambda D_i (1 - \bar{M}_i)
\Big]
+
\Psi(\mathbf{A}, \mathcal{G}).
\end{equation} 

Here $D_i$ denotes the duration of segment $s_i$, while $\bar{M}_i$ represents the mean manipulation confidence within the segment. While Eq.~\ref{eq:energy} is written over the entire segment sequence for clarity, inference is updated incrementally online. We enforce this using a fixed-lag commitment strategy: predictions older than few seconds are frozen and cannot be altered by future observations, ensuring the decoding relies only on a bounded temporal buffer. The first term encourages segments to match visually similar action prototypes, with longer segments contributing proportionally to the objective. The second term introduces a visibility prior that penalizes segments lacking reliable interaction evidence. Procedural structure is enforced through the following graph constraint:
\begin{equation}
\Psi(a_{i-1}, a_i) =
\begin{cases}
0 & (a_{i-1}, a_i) \in E_{\mathcal{G}} \\
\infty & \text{otherwise}.
\end{cases}
\end{equation}

\noindent\textbf{Low-Visibility Inertia and Background Assignment.} Egocentric videos often suffer from transient occlusions or rapid head movements where hands momentarily leave the frame. To maintain stability under extremely low hand visibility, we employ a two-part strategy. If the system is actively tracking a procedural step, the decoder applies inertia, automatically extending the previously predicted action with a small constant cost to seamlessly bridge the visual gap. Conversely, if the sequence is just beginning or the system is already in a dynamically suppressed background state, this lack of visibility safely assigns the segment to the background (BG). 
This preserves task continuity during brief occlusions while preventing spurious predictions when no manipulation is visible.

\noindent\textbf{Gap rectification.} 
Egocentric videos often contain short occlusions or detector failures that interrupt otherwise continuous actions. To address these cases, we introduce a rectification mechanism during beam search. If a background segment is followed by a high-confidence return to the preceding action, the decoder retroactively assigns that action label to the gap. The rectified energy becomes
\begin{equation}
E_\text{rectified} =
E_\text{prev}
-
\text{Cost}_{BG}
+
(D_\text{gap} \cdot \bar{\phi}_{a_{i-2}}).
\end{equation}

This mechanism allows the model to bridge short visual interruptions while maintaining procedural consistency.

\section{Experiments}
\label{sec:experiments}

\subsection{Datasets and Metrics}

We evaluate our framework on two benchmark datasets for egocentric procedural activity: GTEA~\cite{fathi2011learning} and EgoPER~\cite{lee2024error}. 
GTEA consists of 28 videos spanning 7 fine-grained kitchen activities, which we process at a temporal resolution of 15 fps; results on GTEA are averaged over its four standard splits.
EgoPER contains 216 normal and 173 erroneous egocentric videos across 5 recipes; following~\cite{shen2024progress}, we evaluate only on the normal videos and use the same data splits. For EgoPER we operate at a temporal resolution of 10 fps. We report standard procedure parsing metrics including frame-wise Accuracy (Acc) without background frames, Edit distance, and F1 scores at overlap thresholds $\{0.1, 0.25, 0.5\}$ to capture both classification accuracy and temporal alignment quality. However, these standard segmentation metrics primarily measure local boundary accuracy and do not capture whether predicted action sequences preserve long-range procedural dependencies.
To address this limitation, we additionally introduce \textit{N-Step Transition Accuracy} to evaluate structured parsing.

This metric extracts all $n$-step action transitions from the predicted sequence and compares them with the ground-truth to measure consistency of multi-step dependencies. 
We compute precision–recall curves for these transitions and report the area under the curve (AUC). To evaluate performance under partial observations, we sweep the evaluation over progressive video completion levels. 
Specifically, each video is parsed using only the first $\{10\%,20\%,\dots,100\%\}$ of its frames. 
This setting quantifies a model's ability to maintain a coherent procedural state as additional evidence becomes available.

\subsection{Implementation Details}
\label{subsec:impl_dets}

Our framework utilizes a frozen ViT-L/14 DINOv2 backbone~\cite{oquab2023dinov2} as the feature extractor. We apply a frozen hand-object detector (HOD) from~\cite{shan2020understanding} to obtain bounding boxes for hands and active objects in each frame.
\textit{Manipulation-Anchored Features.} We compute the minimum spanning box covering these detections to define the interaction region. We then identify the $16\times16$ DINOv2 patch tokens that intersect with this region. The final frame representation $\mathbf{f}_t$ is obtained by averaging the selected tokens.
\textit{Training-Free Boundary Detection.} The checkerboard kernel width $L$ is set to 20 frames for EgoPER and 15 frames for GTEA.
\textit{Prototype Matching.} The number of clusters is set to $k=3$ for GTEA and $k=9$ for EgoPER. To construct the micro-prototypes, we temporally slice these cluster representations using a segment length of 4 seconds with a 2-second stride for EgoPER, and a 1.5-second segment length with a 0.5-second stride for GTEA. \textit{Graph-Constrained Procedural Inference.} Beam search decoding uses beam width $B=10$ for both datasets.

\begin{table*}[t]
\setlength{\tabcolsep}{8pt} %
    \centering
    \caption{\textbf{Action Segmentation on Procedural Datasets}: Comparison to offline/online baselines on GTEA and EgoPer. We achieve SoTA result on all metrics.}
    \resizebox{\textwidth}{!}{%
    \begin{tabular}{>{\kern-\tabcolsep}lcccccccccccc<{\kern-\tabcolsep}}
        \toprule
         &  & &  \multicolumn{5}{c}{\textbf{GTEA}} & \multicolumn{5}{c}{\textbf{EgoPer}} \\
        \cmidrule(lr){4-8} \cmidrule(lr){9-13} %
         \textbf{Method}& \textbf{Inference} & \textbf{Train-Free?} & Acc & Edit & \multicolumn{3}{c}{F1@\{0.1,0.25,0.5\}} & Acc & Edit & \multicolumn{3}{c}{F1@\{0.1,0.25,0.5\}} \\
        \midrule
        
        MSTCN~\cite{sekaran2022mstcn} & Offline &  \XSolidBrush & 79.35 & 84.46 & 86.54 & 83.79 & 71.86 & 
        87.52 & 92.34 & 92.60 &  91.81 & 86.12 \\
        \midrule
        
        MSTCN & Online & \XSolidBrush & 47.28 & 60.26 & 66.75 & 60.12 & 40.29 & 
        25.40 & 44.81 & 44.09 &  31.78 & 15.72 \\
        
        ProTAS~\cite{shen2024progress} & Online & \XSolidBrush & 73.19 & 71.81 & 72.89 & 68.94 & 54.87 & 
        76.61 & 65.50 & 64.26 & 62.59 & 51.31 \\
        
        \rowcolor{blue!10}
        Ours & Online & \Checkmark & \textbf{82.7} & \textbf{89.3} & \textbf{87.6} & \textbf{85.1} & \textbf{74.0}
        & \textbf{80.7} & \textbf{88.7} & \textbf{91.1} & \textbf{88.5}& \textbf{77.6} \\
        \bottomrule
    \end{tabular}
    }
    \label{tab:performance_comparison}
    
\end{table*}

\section{Results}
\label{sec:results}

Despite requiring no training or fine-tuning, VidParse achieves competitive performance with strong online baselines while maintaining improved procedural consistency. We evaluate our method on several axes. First, We analyze action segmentation performance on standard benchmarks. We then study procedural parsing accuracy to access whether predicted sequences preserve long-range task structure. Next, we examine the temporal behavior of our online boundary detection mechanism. Finally, we present ablation studies to better understand the design choices of our framework.

\subsection{Action Segmentation Performance}

We evaluate action segmentation on the GTEA and EgoPER datasets. Tab.~\ref{tab:performance_comparison} reports results using Accuracy, Edit distance, and F1 scores at multiple overlap thresholds. Our method consistently outperforms the ProTAS baseline, improving F1 scores by $15$--$19\%$ across thresholds on GTEA, with the largest gain on F1@0.5. On EgoPER, we observe improvements of roughly $25\%$ on F1 and Edit metrics while maintaining a $4\%$ gain in frame-level accuracy.

\begin{figure}[t]
    \centering
    \includegraphics[width=\linewidth]{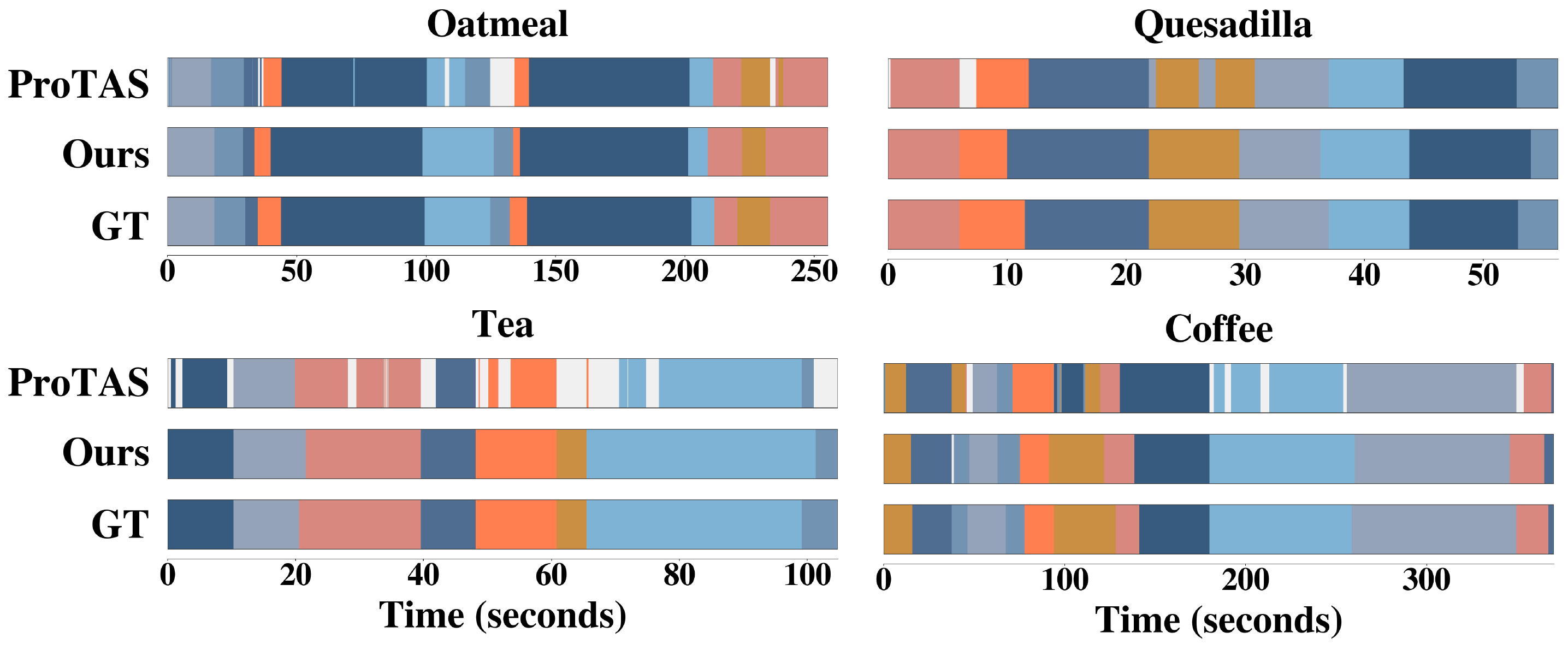}
    \caption{\textbf{Action Segmentation Performance}: Frame-level action segmentation models, such as ProTAS are brittle and result in inconsistent parsing. Our structured procedural inference results in smooth and stable action transitions. (Background GT frames have been removed here for brevity)}
    \label{fig:action_seg}
\end{figure}

Unlike frame-level models that predict labels independently for each frame, our approach produces temporally coherent action segments. As a result, small boundary offsets may slightly affect frame accuracy even when predicted segments align well with the ground-truth. Metrics such as Edit distance and F1@k therefore better reflect segmentation quality in this setting. As illustrated in Fig.~\ref{fig:action_seg}, our predictions remain smooth and closely follow the ground-truth segmentation without the flickering commonly observed in frame-level models.

\subsection{Procedural Parsing Accuracy}

\begin{figure}[t]
    \centering
    \includegraphics[width=\linewidth]{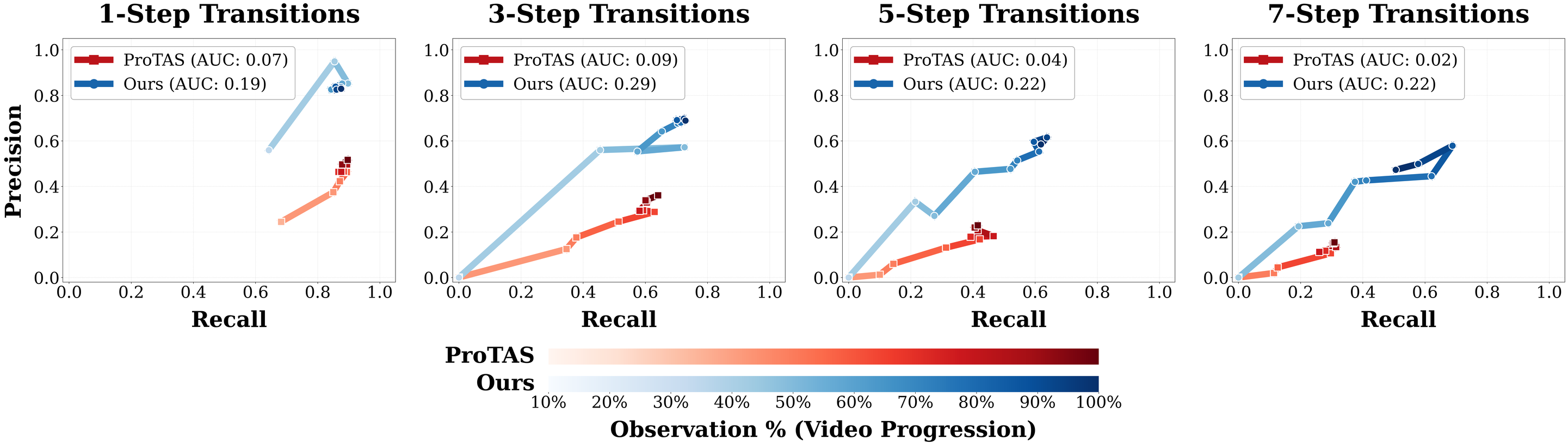}
    \caption{\textbf{N-Step transition performance on EgoPER}: VidParse consistently outperforms the ProTAS baseline, with the gap widening for higher transition lengths.}
    \label{fig:hops}
    
\end{figure}

Procedural tasks require maintaining a consistent sequence of actions rather than recognizing actions independently.
To evaluate this property, we measure \emph{N-step Transition Accuracy}, which captures whether predicted sequences preserve valid multi-step transitions. Let $\mathcal{H}_n(S)$ denote the \textit{multiset} of all $n$-step transitions in sequence $S$:
\begin{equation}
    \mathcal{H}_n(S) = \{ (s_i, s_{i+1}, \dots, s_{i+n}) \mid 1 \leq i \leq k-n \}
\end{equation}
We evaluate performance by comparing the predicted multiset $\mathcal{H}_n(S_{\text{pred}})$ 
with the ground-truth multiset $\mathcal{H}_n(S_{\text{gt}})$. 
Let $\mathcal{U}$ be the set of unique transition tuples present in either sequence. 
For any transition tuple $h \in \mathcal{U}$, let $c(h, S)$ denote the count (multiplicity) 
of $h$ in $\mathcal{H}_n(S)$. True Positive ($TP$) count is defined as:
\begin{equation}
    TP_n = \sum_{h \in \mathcal{U}} \min\left( c(h, S_{\text{pred}}), \ c(h, S_{\text{gt}}) \right)
\end{equation}

As shown in Fig.~\ref{fig:hops}, our method consistently outperforms the ProTAS baseline across all transition lengths. The gap increases with larger $n$: our AUC for 5-step and 7-step transitions is up to $5\times$–$10\times$ higher than the baseline. This indicates that our approach preserves the global task structure more effectively, while frame-level baselines accumulate local errors that disrupt long-range transitions and violate procedural constraints. Our method therefore reconstructs the underlying task graph rather than recognizing actions in isolation.

\subsection{Online Boundary Detection Behavior}

\begin{figure*}[t!] %
    \centering
    \begin{subfigure}{0.45\textwidth}
        \centering
        \includegraphics[width=\linewidth]{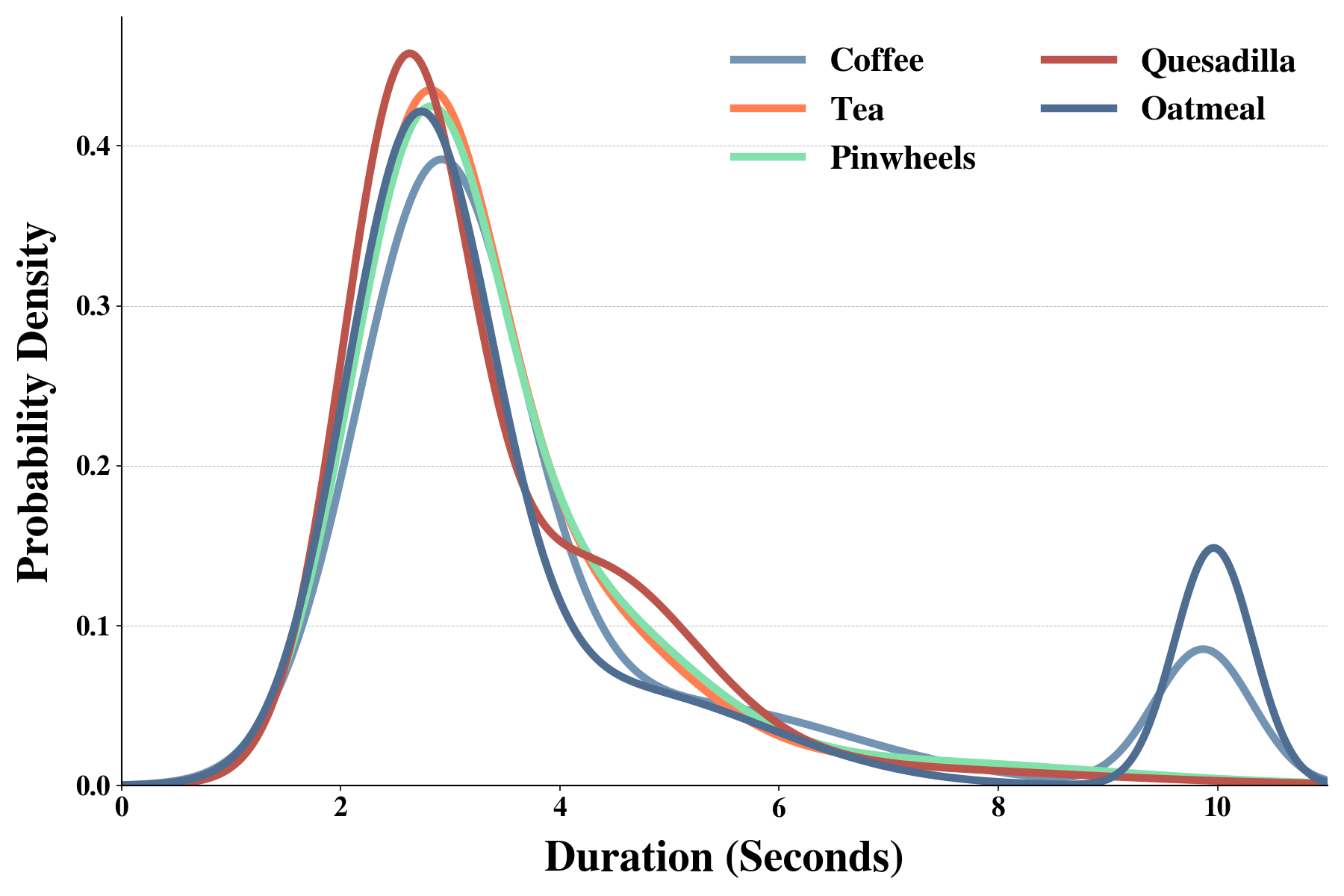}
        \caption{Action duration comparison.}
        \label{fig:duration_comp}
    \end{subfigure}
    \hfill %
    \begin{subfigure}{0.45\textwidth}
        \centering
        \includegraphics[width=\linewidth]{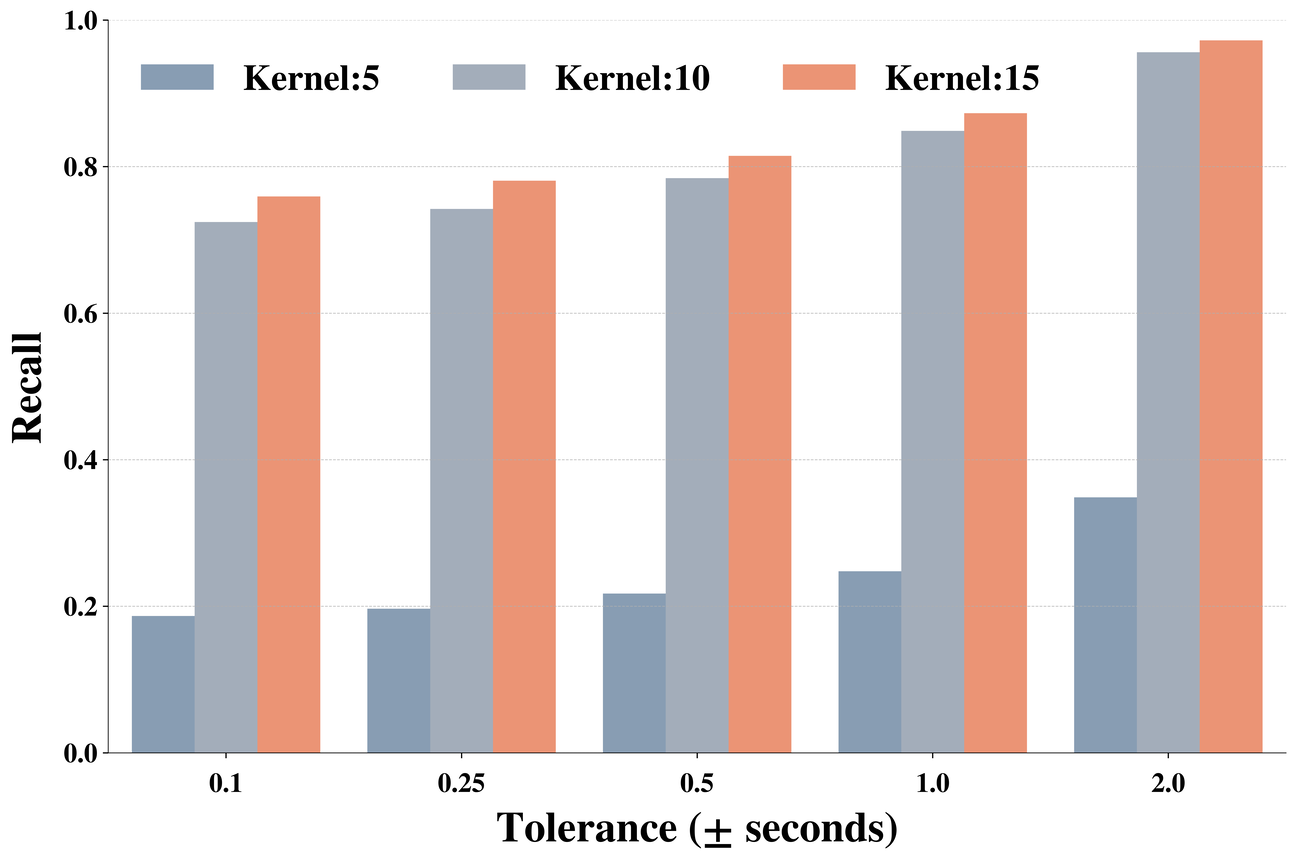}
        \caption{Boundary recall analysis.}
        \label{fig:recall_plot}
    \end{subfigure}

    \caption{\textbf{Training-Free Boundary Analysis:} (a) 75\% of inference segments are under 4s showing our online nature (b) at $kernel{=}10$, we can recover a boundary within 2\,s over 90\% of the time showing that we do not mix actions}
    \label{fig:combined_boundary_analysis}
    
\end{figure*}

We analyze the temporal behavior of our online boundary detection mechanism. Fig.~\ref{fig:combined_boundary_analysis}(a) shows the distribution of emitted segment durations. Approximately $75\%$ of predicted segments fall within a 2–4 second window, and more than $90\%$ are shorter than 5 seconds. We compare the performance of our boundary detection method with other online unsupervised methods in  Tab.~\ref{tab:combined_ablations}(b) Bottom-Right. \textit{Adjacent Frame Similarity} emits a boundary when the cosine distance between consecutive frame embeddings exceeds a fixed, predefined threshold. \textit{z-score} triggers a boundary when the frame-to-frame distance deviates from the local mean and standard deviation of a rolling temporal window. Our method provides the best performance on the downstream action segmentation metrics.

Operating at the segment level also stabilizes predictions over time compared to frame-level methods. Frame-level models often produce rapid label fluctuations near action boundaries, whereas our method groups semantically consistent frames into discrete segments. As illustrated in Fig.~\ref{fig:action_seg}, the resulting predictions closely match the ground-truth structure and remain temporally stable.

\subsection{Ablation Studies}

\noindent\textbf{Feature Representation.}
We analyze the utility of using MAF representations  in Tab.~\ref{tab:feature_ablation}, and further compare them against egocentric video-language backbone, EgoVLPv2 \cite{pramanick2023egovlpv2} in Tab.~\ref{tab:feature_ablations} (Right). Starting from the ProTAS baseline, incorporating MAF features leads to consistent improvements across segmentation metrics, indicating that manipulation-anchored representations provide stronger cues for procedural actions. Applying our method on top of the same MAF features further improves performance across all metrics and achieves the best overall results. Full-frame features underperform MAFs, suggesting that explicit foreground information is necessary. Frame-level EgoVLPv2 remains competitive with DINOv2 MAFs, while clip-level features perform worse, indicating that coarse temporal representations can miss short procedural transitions. 

\noindent\textbf{Prototype Method.} We evaluate prototype methods in Tab.~\ref{tab:combined_ablations}(a). While computing a global mean collapses the action feature space, relying strictly on medoids limits the representation to previously observed instances. Our centroid-based approach better captures the action feature space while remaining representative of the underlying prototypes. This strategy yields the best performance across all metrics, reaching an Acc. of 80.69 and an F1@0.5 of 77.61. Further, Tab.~\ref{tab:combined_ablations}(d) demonstrates that our clustering approach for action matching is  robust. Varying the number of prototypes $p$, extracted per action class, from 3 to 11 results in less than a 1.2$\%$ variance in Accuracy and Edit score.

\noindent\textbf{Inference Speed.} In Tab.~\ref{tab:feature_ablations} Left, we analyze the impact of beam width ($B$) on inference speed. Although MAF extraction adds HOD overhead, our approach is significantly faster, achieving 8.6 FPS, than the ProTAS baseline (4.5 FPS), allowing for a highly efficient and tunable speed-accuracy trade-off. Also, our parsing and decoding stages, operating on an AMD EPYC 7443 CPU, require only 32.8s vs 706s needed by ProTAS running on an NVIDIA RTX A4000.

\noindent\textbf{Minium distance \textit{d}} As shown in Table~\ref{tab:combined_ablations}(c), the minimum gap parameter $d$ controls how close consecutive boundaries can be. Smaller values allow denser boundaries and can mildly over-segment actions, while larger values impose rigid spacing and may miss rapid transitions.

\begin{table*}[t]
    \centering
    \caption{\textbf{MAFs vs other features}: MAFs outperform other features across methods on EgoPER}
    \label{tab:feature_ablation}
    \setlength{\tabcolsep}{2.5pt} %
    \resizebox{\textwidth}{!}{%
    \begin{tabular}{ll c ccc ccc ccc ccc ccc ccc}
        \toprule
        & & & \multicolumn{3}{c}{\textbf{Coffee}} & \multicolumn{3}{c}{\textbf{Oatmeal}} & \multicolumn{3}{c}{\textbf{Pinwheels}} & \multicolumn{3}{c}{\textbf{Quesdilla}} & \multicolumn{3}{c}{\textbf{Tea}} & \multicolumn{3}{c}{\textbf{Overall}} \\
        \cmidrule(lr){4-6} \cmidrule(lr){7-9} \cmidrule(lr){10-12} \cmidrule(lr){13-15} \cmidrule(lr){16-18} \cmidrule(l){19-21}
        \textbf{Method} & \textbf{Feat.} & \textbf{Train-Free?} & Acc & Edit & F1 & Acc & Edit & F1 & Acc & Edit & F1 & Acc & Edit & F1 & Acc & Edit & F1 & Acc & Edit & F1 \\
        \midrule
        ProTAS~\cite{shen2024progress} & I3D &  \XSolidBrush & 60.58 & 36.75 & 24.81 & 86.84 & 76.33 & 65.85 & 78.74 & 75.82 & 55.68 & 78.98 & 62.18 & 53.81 & 77.94 & 76.44 & 56.42 & 76.61 & 65.50 & 51.31 \\
        ProTAS & DINO CLS &  \XSolidBrush & 66.42 & \textbf{50.40} & 39.16 & \textbf{88.85} & \textbf{77.22} & \textbf{74.55} & 82.33 & 81.54 & 72.34 & 83.58 & 74.80 & 67.48 & 85.37 & \textbf{79.18} & 68.68 & 81.31 & \textbf{72.63} & 64.44 \\
        ProTAS & DINO MAFs & \XSolidBrush & \textbf{71.80} & 49.17 & \textbf{46.04} & 87.47 & 70.67 & 70.16 & \textbf{83.28} & \textbf{84.39} & \textbf{74.62} & \textbf{86.71} & \textbf{75.37} & \textbf{71.83} & \textbf{87.28} & 74.75 & \textbf{73.64} & \textbf{83.31} & 70.87 & \textbf{67.26} \\
        \midrule
        Ours & DINO CLS & \Checkmark & 65.06 & 74.10 & 58.66 & 74.74 & 86.11 & 60.54 & 69.38 & \textbf{86.25} & 56.06 & 69.68 & 86.92 & 65.41 & 65.77 & 78.04 & 64.52 & 69.09 & 82.68 & 61.42 \\
        \rowcolor{blue!10}
        Ours & DINO MAFs & \Checkmark & \textbf{75.82} & \textbf{76.45} & \textbf{73.37} & \textbf{80.40} & \textbf{91.57} & \textbf{69.70} & \textbf{74.58} & 81.78 & \textbf{57.94} & \textbf{83.58} & \textbf{91.90} & \textbf{89.71} & \textbf{86.42} & \textbf{96.74} & \textbf{91.97} & \textbf{80.69} & \textbf{88.69} & \textbf{77.61} \\
        \bottomrule
    \end{tabular}
    }
\end{table*}

\begin{table}[t]
\centering
\caption{\textbf{Inference Speed and Feature Comparisons on EgoPER.} \textbf{(Left)} Impact of beam width ($B$) on inference speed and parsing accuracy. The times shown are average seconds/video. Our CPU-based method is faster than the GPU-accelerated ProTAS baseline \textbf{(Right)} Feature comparisons highlighting the importance of frame-level MAFs over clip-level or full-frame features.}
\label{tab:feature_ablations}

\renewcommand{\arraystretch}{1.05} 

\begin{minipage}[t]{0.67\textwidth}
  \begin{adjustbox}{max width=\textwidth, valign=t}
  \begin{tabular}{lcccccccc}
    \toprule
    \textbf{Set} & \textbf{HOD} & \textbf{Backbone} & \textbf{Method} & \textbf{Total} & \textbf{FPS} & \textbf{Acc} & \textbf{Edit} & \textbf{F1@0.5} \\
    \midrule
    ProTAS~\cite{shen2024progress} & - & 79.5 & 706 & 785.5 & 4.5 & 76.6 & 65.5 & 51.3 \\
    \midrule
    Ours ($B=1$) & 339 & 36.8 & \textbf{24.8} & \textbf{400.6} & \textbf{8.7} & 58.5 & 77.5 & 52.1 \\
    Ours ($B=3$) & 339 & 36.8 & 28.2 & 404.0 & \textbf{8.7} & 76.3 & 86.0 & 72.4 \\
    Ours ($B=5$) & 339 & 36.8 & 28.4 & 404.2 & 8.6 & 78.5 & 86.9 & 74.5 \\
    \rowcolor{blue!10}
    Ours ($B=10$) & 339 & 36.8 & 32.8 & 408.6 & 8.6 & \textbf{80.7} & \textbf{88.7} & \textbf{77.6} \\
    \bottomrule
  \end{tabular}
  \end{adjustbox}
\end{minipage}\hfill
\begin{minipage}[t]{0.30\textwidth}
  \begin{adjustbox}{width=\textwidth, valign=t}
  \begin{tabular}{lccccc}
    \toprule
    \textbf{Method} & \textbf{MAF} & \textbf{Clip} & \textbf{Acc} & \textbf{Edit} & \textbf{F1} \\
    \midrule
    EgoVLP2 & \XSolidBrush & \XSolidBrush & 78.1 & 89.4 & 73.0 \\
    EgoVLP2 & \Checkmark   & \XSolidBrush & 80.3 & \textbf{91.5} & 77.5 \\
    EgoVLP2 & \XSolidBrush & \Checkmark   & 53.4 & 62.8 & 43.7 \\
    EgoVLP2 & \Checkmark   & \Checkmark   & 60.1 & 70.6 & 52.1 \\
    Ours    & \XSolidBrush & \XSolidBrush & 69.1 & 82.7 & 61.4 \\
    \rowcolor{blue!10}
    Ours    & \Checkmark   & \XSolidBrush & \textbf{80.7} & 88.7 & \textbf{77.6} \\
    \bottomrule
  \end{tabular}
  \end{adjustbox}
\end{minipage}

\end{table}

\begin{table}[t!]
  \centering
  \caption{\textbf{Ablation and Hyperparameter Sensitivity Analyses on EgoPER}(a) prototype matching strategies and (b) boundary detection methods, alongside sensitivity to (c) the gap threshold $d$ and (d) the number of prototypes $p$.}
  \label{tab:combined_ablations}
  \setlength{\tabcolsep}{1.2pt} %
  \renewcommand{\arraystretch}{0.9}

  \begin{minipage}[t]{0.33\columnwidth}
    \centering
    {\scriptsize \textbf{(a)} Matching}\\
    \resizebox{\linewidth}{!}{%
    \begin{tabular}{lccc}
      \toprule
      Type & Acc & Edit & F1 \\
      \midrule
      Global & 78.5 & 88.8 & 74.5 \\
      Medoids & 80.1 & 87.0 & 75.0 \\
      \rowcolor{blue!10}
      Centroids & \textbf{80.7} & \textbf{88.7} & \textbf{77.6} \\
      \bottomrule
    \end{tabular}}
  \end{minipage}
  \hfill
  \begin{minipage}[t]{0.29\columnwidth}
    \centering
    {\scriptsize \textbf{(b)}Boundary}\\
    \resizebox{\linewidth}{!}{%
    \begin{tabular}{lccc}
      \toprule
      Method & Acc & Edit & F1 \\
      \midrule
      Adj. Frame & 75.7 & 83.4 & 68.0 \\
      Z-score & 78.5 & 85.1 & 69.9 \\
      ABD~\cite{du2022fast} & 78.5 & 88.1 & 72.2 \\
      \rowcolor{blue!10}
      Ours & \textbf{80.7} & \textbf{88.7} & \textbf{77.6} \\
      \bottomrule
    \end{tabular}}
  \end{minipage}
  \hfill
  \begin{minipage}[t]{0.17\columnwidth}
    \centering
    {\scriptsize \textbf{(c) $d$}}\\
    \resizebox{\linewidth}{!}{%
    \begin{tabular}{cccc}
      \toprule
      $d$ & Acc & Edit & F1 \\
      \midrule
      10 & 81.0 & 84.1 & 73.5 \\
      15 & \textbf{81.7} & 87.9 & 75.7 \\
      \rowcolor{blue!10}
      20 & 80.7 & \textbf{88.7} & \textbf{77.6} \\
      25 & 78.1 & 87.7 & 72.8 \\
      30 & 76.1 & 87.7 & 71.3 \\
      \bottomrule
    \end{tabular}}
  \end{minipage}
  \hfill
  \begin{minipage}[t]{0.17\columnwidth}
    \centering
    {\scriptsize \textbf{(d) $p$}}\\
    \resizebox{\linewidth}{!}{%
    \begin{tabular}{cccc}
      \toprule
      $p$ & Acc & Edit & F1 \\
      \midrule
      3  & 80.5 & 89.0 & 75.8 \\
      5  & 79.7 & 89.2 & 76.0 \\
      7  & 80.1 & \textbf{89.9} & 76.5 \\
      \rowcolor{blue!10}
      9  & \textbf{80.7} & 88.7 & \textbf{77.6} \\
      11 & 80.4 & 89.2 & 77.1 \\
      \bottomrule
    \end{tabular}}
  \end{minipage}
  
\end{table}

\section{Limitations}
Our causal framework does not require global context, but its segment-level processing may be unsuitable for ultra-low-latency applications. Its learned task graph may also generalize poorly to unseen procedural variations. In addition, the Hand-Object Detector (HOD) can produce incorrect or delayed predictions under severe occlusion, motion blur, or non-interactive actions (e.g., “Using a Microwave”). The framework may extend to exocentric videos when manipulation cues are visible, but performance may degrade when they are weak/absent. Future work could use probabilistic graphs to better handle open-world and loosely structured tasks. See the supplementary material and project page for details.

\section{Conclusion}
We presented VidParse, an online, training-free framework for egocentric procedure parsing that treats activity understanding as a structured inference problem. By leveraging Manipulation-Anchored Features and a Gaussian-tapered checkerboard kernel applied to frozen DINOv2 representations, our method detects action boundaries directly from visual evidence without gradient-based learning. We further integrate task-graph constrained beam search with gap rectification to enforce procedural consistency and correct transient visual failures. Experiments demonstrate that combining structural priors with foundation model features yields competitive performance with strong online baselines while better preserving long-range procedural structure.

\section{Acknowledgments}
This work was partially supported by NSF CAREER Award (\#2238769) to Abhinav Shrivastava. The authors acknowledge UMD's supercomputing resources made available for conducting
this research. The U.S. Government is authorized to reproduce and distribute reprints for Governmental purposes notwithstanding any copyright annotation thereon. The views and conclusions contained herein are those of the authors and should not be interpreted as necessarily representing the official policies or endorsements, either expressed or implied, of NSF or the U.S. Government.

\bibliographystyle{splncs04}
\bibliography{references}

\clearpage

\renewcommand{\theHsection}{S\arabic{section}}
\renewcommand{\theHfigure}{S\arabic{figure}}
\renewcommand{\theHtable}{S\arabic{table}}
\renewcommand{\theHequation}{S\arabic{equation}}

\addtocontents{toc}{\protect\setcounter{tocdepth}{2}}
\pdfbookmark[0]{Supplementary Material}{suppltitle}

\title{Supplementary Material for:\\ VidParse: Online Parsing of Egocentric Procedures Like a Pro} 
\titlerunning{Supplementary Material for VidParse}

\author{Anubhav Gupta \and
Archit Kambhamettu \and
Vatsal Agarwal \and
Pulkit Kumar \and
Abhinav Shrivastava}

\authorrunning{A.~Gupta et al.}

\institute{University of Maryland, College Park, USA\\
\email{\{anubhav, architk, vatsalag, pulkit, abhinav2\}@umd.edu}}

{
\renewcommand{\addcontentsline}[3]{}
\maketitle
}
\setcounter{tocdepth}{2} %
\begingroup
\let\clearpage\relax
\tableofcontents
\endgroup
\vspace{2em}

\setcounter{section}{0}
\renewcommand{\thesection}{\arabic{section}}
\setcounter{figure}{0}
\renewcommand{\thefigure}{\arabic{figure}}
\setcounter{table}{0}
\renewcommand{\thetable}{\arabic{table}}
\setcounter{equation}{0}
\renewcommand{\theequation}{\arabic{equation}}

\section{Qualitative Structured Parsing Results}
To better illustrate the temporal stability and structural accuracy of our approach, we provide interactive qualitative results in the attached \href{run:./index.html}{index.html} file. This supplementary web page includes complete video parses for various recipes (Coffee, Tea, Oatmeal, Quesadilla, and Pinwheels) in EgoPer. For each sequence, viewers can observe the real-time action segmentation alongside a synchronized visualization of the underlying task graph, demonstrating how VidParse prunes invalid transitions and maintains a coherent multi-step trajectory even amidst egocentric noise.

\section{Task Graphs}

\begin{figure*}[t!] %
    \centering
    \begin{subfigure}{0.25\textwidth}
        \centering
        \includegraphics[width=\linewidth]{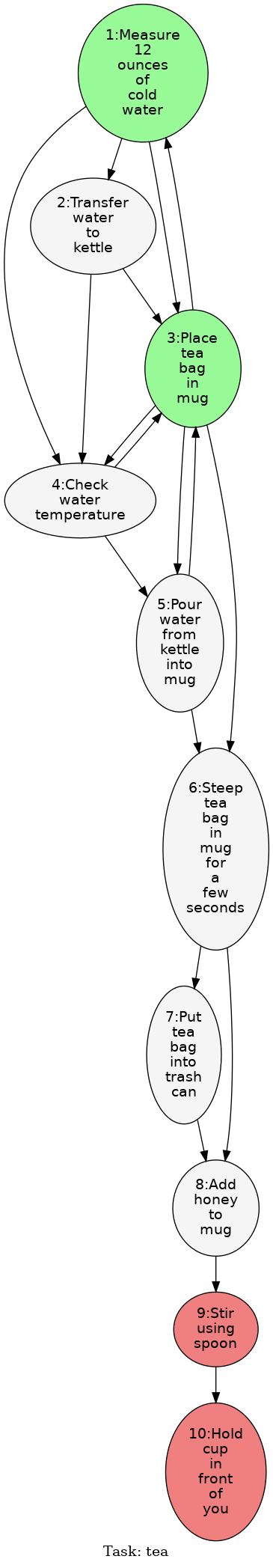}
        \caption{Tea}
        \label{fig:Tea_graph}
    \end{subfigure}
    \begin{subfigure}{0.25\textwidth}
        \centering
        \includegraphics[width=\linewidth]{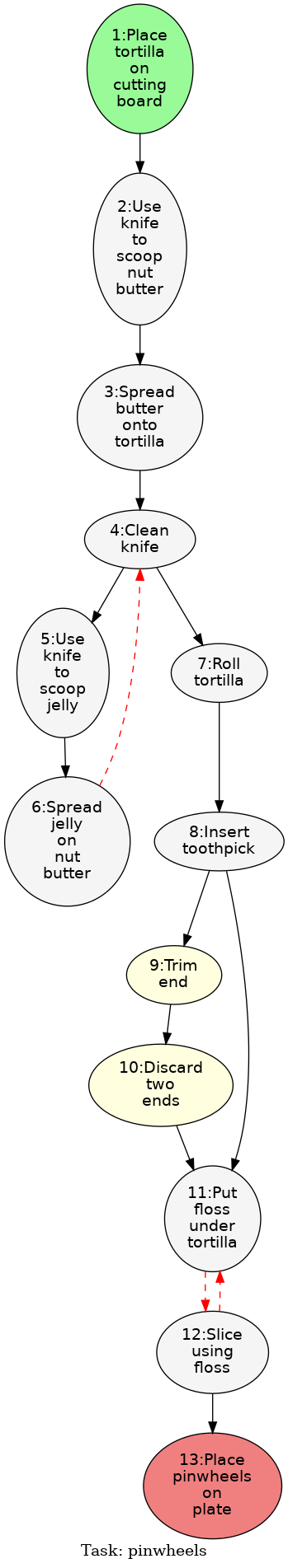}
        \caption{Pinwheels}
        \label{fig:pinwheels_graph}
    \end{subfigure}
    \begin{subfigure}{0.45\textwidth}
        \centering
        \includegraphics[width=\linewidth]{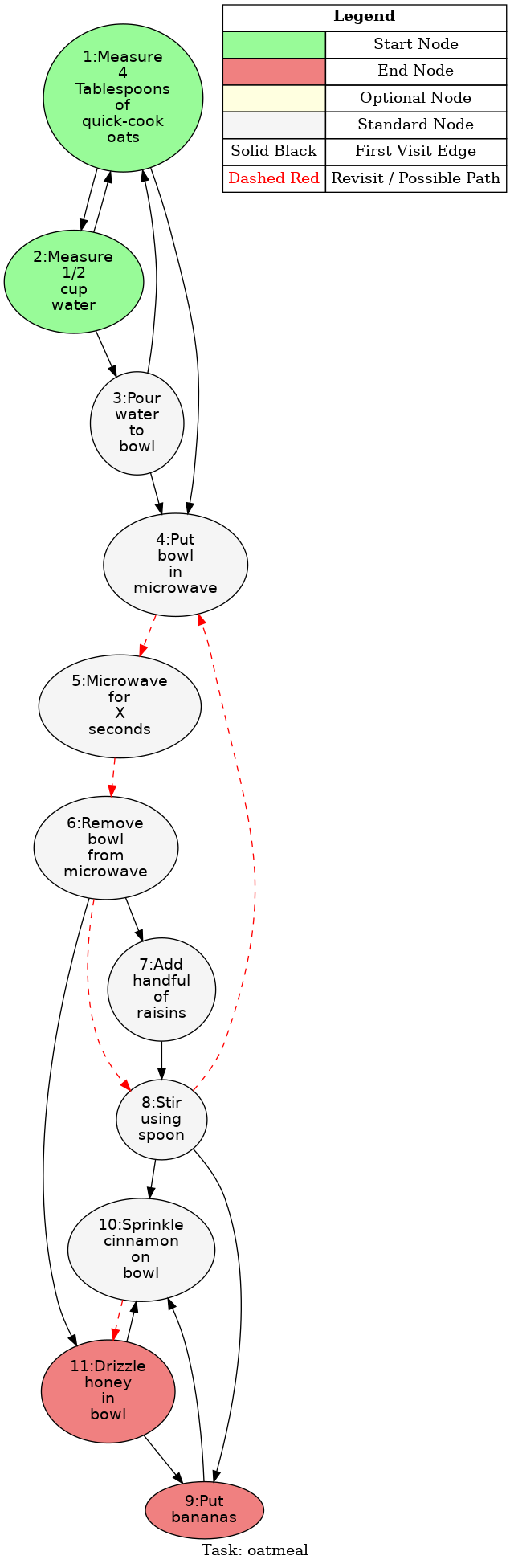}
        \caption{Oatmeal}
        \label{fig:oatmeal_graph}
    \end{subfigure}

    \caption{\textbf{Parsed Task Graphs on EgoPer:} (a) Tea (b) Pinwheels (c) Oatmeal}
    \label{fig:taskgraphs_a}
    \vspace{-1em} %
\end{figure*}

\begin{figure*}[t!] %
    \centering
    \begin{subfigure}{0.45\textwidth}
        \centering
        \includegraphics[width=\linewidth]{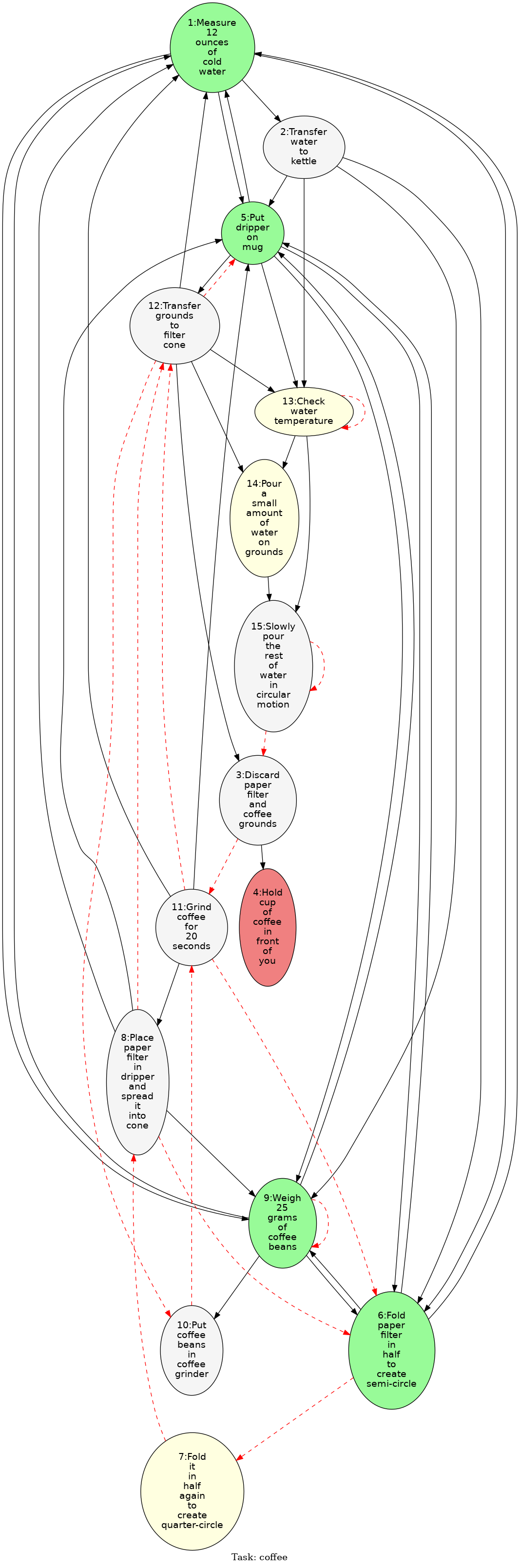}
        \caption{Coffee}
        \label{fig:coffee_graph}
    \end{subfigure}
    \begin{subfigure}{0.45\textwidth}
        \centering
        \includegraphics[width=\linewidth]{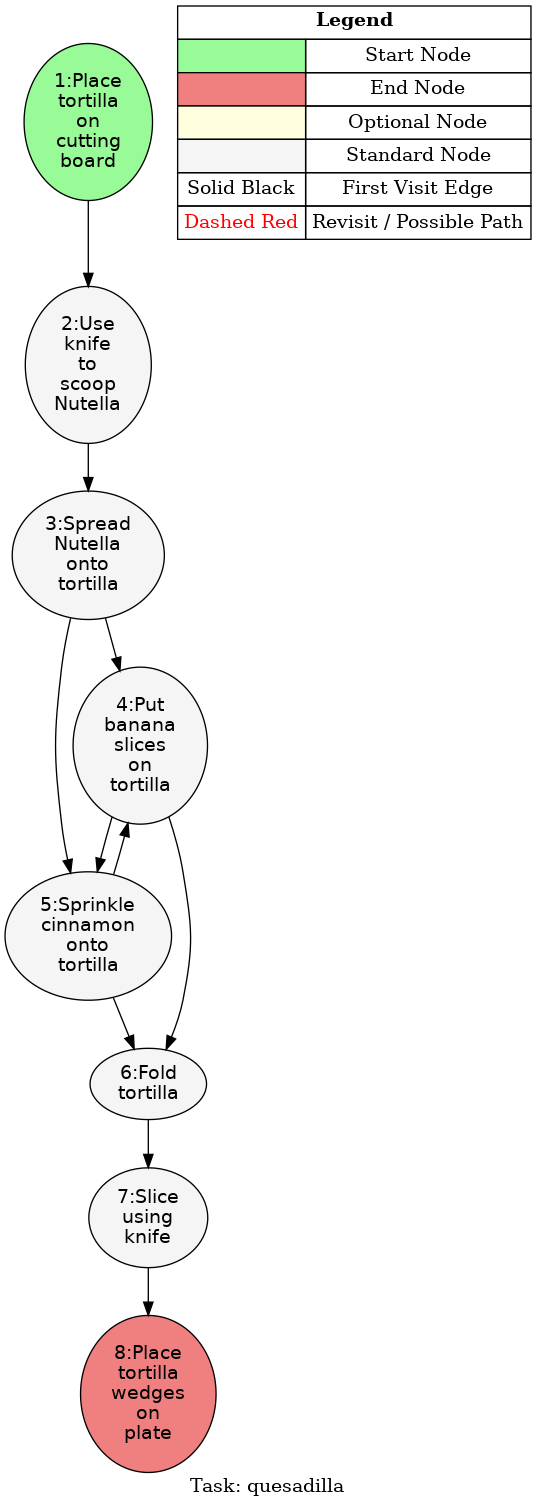}
        \caption{Quesadilla}
        \label{fig:quesadilla_graph}
    \end{subfigure}

    \caption{\textbf{Parsed Task Graphs on EgoPer:} (a) Coffee (b) Quesadilla}
    \label{fig:taskgraphs_b}
    \vspace{-1em} %
\end{figure*}

To model the temporal structure and inherent variability of complex activities, we construct a directed graph $G = (V, E)$ for each recipe by aggregating temporal action sequences across all videos demonstrating that specific task. Each node $v \in V$ represents a distinct atomic action or step, while each directed edge $e = (u, v) \in E$ represents an observed sequential transition from action $u$ to action $v$. Because egocentric videos often contain significant behavioral variance—such as skipped steps, repeated actions, and differing end states—the graphs in Fig.~\ref{fig:taskgraphs_a} and Fig.~\ref{fig:taskgraphs_b} visualize these structural nuances using specific node colors and edge styles.

\textbf{Node Definitions (Actions/Steps)}:
\begin{itemize}
    \item Green Nodes (Start Nodes): Represent initiating actions. These are steps that can be performed without any prior prerequisite actions having been completed.
    \item Red Nodes (End Nodes): Represent terminal actions. A node is marked as an end node if it serves as the final, concluding step in at least one video sequence for that recipe.
    \item Yellow Nodes (Optional Nodes): Represent omittable actions. These are steps that appear in some, but not all, video sequences for the given recipe, indicating that the task can be successfully completed without them.
    \item Gray/White Nodes (Standard Nodes): Represent mandatory intermediate actions that are present in all observed video sequences for the recipe.
\end{itemize}

\textbf{Edge Definitions (Transitions)}:
\begin{itemize}
    \item Solid Black Edges (First-Visit Paths): Represent the primary sequential flow. These edges indicate the first time a transition to a new action is observed within a video sequence.
    \item Dashed Red Edges (Revisit Paths): Represent cyclical or repeated behavior. These edges indicate a transition to an action that has already been executed previously within the same video sequence, highlighting corrections, looping steps, or alternative chronological pathways.
\end{itemize}

\section{Feature Representation Ablation on GTEA}
In the main text, we demonstrated the efficacy of VidParse against strong online baselines. In some generalized settings, procedural parsing is modeled primarily as an action graph, aggregating atomic verbs (e.g., take, pour, place) across various recipes without strict object dependence. While this generalized action graph captures broad temporal dynamics, our focus is on exact procedural execution, which relies heavily on recipe-specific task graphs. A true procedural task graph inherently couples the action with the specific object being manipulated (e.g., take bread versus take cup), enforcing precise sequential constraints for a given recipe.

To provide a comprehensive analysis, Table~\ref{tab:feature_ablation_supple} evaluates both our framework and the ProTAS baseline under this more rigorous, recipe-specific task graph setting on the GTEA dataset. By isolating each recipe (e.g., Cheese, Coffee, Hotdog), we assess how well different feature representations support fine-grained, object-aware procedural reasoning.

As shown in the table, Manipulation-Anchored Features (MAFs) consistently outperform other representations. Because MAFs are explicitly designed to capture hand-object interactions, they naturally excel in a task graph setting where precise object context is required to progress through the procedural steps. Even when the ProTAS baseline is adapted to utilize this recipe-specific structure alongside MAF representations, it plateaus at an overall F1 score of 21.30. In contrast, VidParse leverages the same MAFs within our strict graph-constrained inference to achieve an overall F1 of 74.03. Notably, on specific recipes like ``Cheese'' our approach achieves near perfect structural parsing. This underscores that while strong features are beneficial, our training-free, structurally constrained decoding is uniquely suited for exact procedural understanding over complex, recipe-specific trajectories.

\begin{table*}[t]
    \centering
    \caption{\textbf{MAFs vs other features}: MAFs outperform other features across methods on GTEA. ${}^{*}$ indicates our recipe-specific task-graph setting.}
    \label{tab:feature_ablation_supple}
    \vspace{-0.1in}
    \setlength{\tabcolsep}{2.5pt} %
    \resizebox{\textwidth}{!}{%
    \begin{tabular}{ll c ccc ccc ccc ccc ccc ccc ccc ccc} %
        \toprule
        & & & \multicolumn{3}{c}{\textbf{Cheese}} & \multicolumn{3}{c}{\textbf{Coffee}} & \multicolumn{3}{c}{\textbf{CofHoney}} & \multicolumn{3}{c}{\textbf{Hotdog}} & \multicolumn{3}{c}{\textbf{Pealate}} & \multicolumn{3}{c}{\textbf{Peanut}} & \multicolumn{3}{c}{\textbf{Tea}} & \multicolumn{3}{c}{\textbf{Overall}} \\
        \cmidrule(lr){4-6} \cmidrule(lr){7-9} \cmidrule(lr){10-12} \cmidrule(lr){13-15} \cmidrule(lr){16-18} \cmidrule(lr){19-21} \cmidrule(lr){22-24} \cmidrule(l){25-27} 
        \textbf{Method} & \textbf{Feat.} & \textbf{Train-Free?} & Acc & Edit & F1 & Acc & Edit & F1 & Acc & Edit & F1 & Acc & Edit & F1 & Acc & Edit & F1 & Acc & Edit & F1 & Acc & Edit & F1 & Acc & Edit & F1 \\
        \midrule
        ProTAS~\cite{shen2024progress} & I3D &  \XSolidBrush & 79.15 & 82.49 & 68.12 & 71.20 & 64.79 & 49.91 & 84.32 & \textbf{90.91} & 72.98 & 67.40 & 71.47 & 51.94 & 73.26 & 69.56 & 57.21 & 75.27 & 65.50 & 47.55 & 67.14 & 57.98 & 47.99 & 73.96 & 71.81 & 56.53 \\
        ProTAS & I3D* & \XSolidBrush & 58.25 & 65.53 & 37.05 & 34.37 & 24.29 & 8.72 & 54.19 & 53.40 & 25.91 & 35.77 & 47.82 & 15.31 & 50.84 & 39.84 & 26.17 & 42.31 & 45.94 & 17.87 & 47.06 & 49.38 & 20.40 & 46.11 & 46.60 & 21.63 \\
        
        ProTAS & DINO CLS &  \XSolidBrush & 77.95 & 78.93 & 66.49 & 71.22 & 63.40 & 48.60 & 84.05 & 84.65 & 71.91 & 60.07 & 70.28 & 51.64 & 72.10 & 71.42 & \textbf{63.08} & 68.35 & 64.13 & 44.52 & 69.53 & 67.50 & 49.27 & 71.90 & 71.47 & 56.50 \\

        ProTAS & DINO CLS* & \XSolidBrush & 48.13 & 68.98 & 27.21 & 30.54 & 27.98 & 11.58 & 54.99 & 57.14 & 27.23 & 34.82 & 44.98 & 18.86 & 39.86 & 37.44 & 18.34 & 40.38 & 41.72 & 17.72 & 44.00 & 42.97 & 17.32 & 41.82 & 45.89 & 19.75 \\
        
        ProTAS & DINO MAFs & \XSolidBrush & 81.07 & 86.45 & 67.08 & \textbf{74.58} & 74.73 & 57.94 & 84.47 & 88.40 & 74.74 & 68.01 & 66.51 & 50.05 & 74.65 & 75.87 & 55.06 & 68.93 & 67.97 & 46.32 & 69.71 & 67.30 & 48.53 & 74.49 & 75.32 & 57.10 \\

        ProTAS & DINO MAFs* & \XSolidBrush & 52.34 & 60.67 & 30.78 & 38.18 & 34.85 & 12.50 & 55.80 & 49.45 & 28.44 & 24.74 & 35.81 & 10.02 & 47.17 & 42.65 & 23.57 & 49.40 & 45.63 & 24.34 & 46.57 & 42.71 & 19.43 & 44.89 & 44.54 & 21.30 \\
        \midrule
        Ours & DINO CLS* & \Checkmark & 58.41 & 84.38 & 48.32 & 59.34 & 73.16 & 48.06 & 29.66 & 50.00 & 20.91 & 38.70 & 50.00 & 39.55 & 49.08 & 61.20 & 28.34 & 42.70 & 44.15 & 20.19 & 67.07 & 77.24 & 57.74 & 49.28 & 62.88 & 37.59 \\
        \rowcolor{blue!10}
        Ours & DINO MAFs* & \Checkmark & \textbf{90.60} & \textbf{100.00} & \textbf{87.50} & 73.39 & \textbf{77.00} & \textbf{69.79} & \textbf{85.60} & 89.97 & \textbf{79.04} & \textbf{85.86} & \textbf{100.00} & \textbf{78.29} & \textbf{77.39} & \textbf{87.35} & 58.13 & \textbf{77.32} & \textbf{83.13} & \textbf{66.42} & \textbf{88.42} & \textbf{87.84} & \textbf{79.01} & \textbf{82.66} & \textbf{89.33} & \textbf{74.03} \\
        \bottomrule
    \end{tabular}
    }
\end{table*}

\section{Visualizing Boundary Detection}

\begin{figure}[t]
    \centering
    \includegraphics[width=\linewidth]{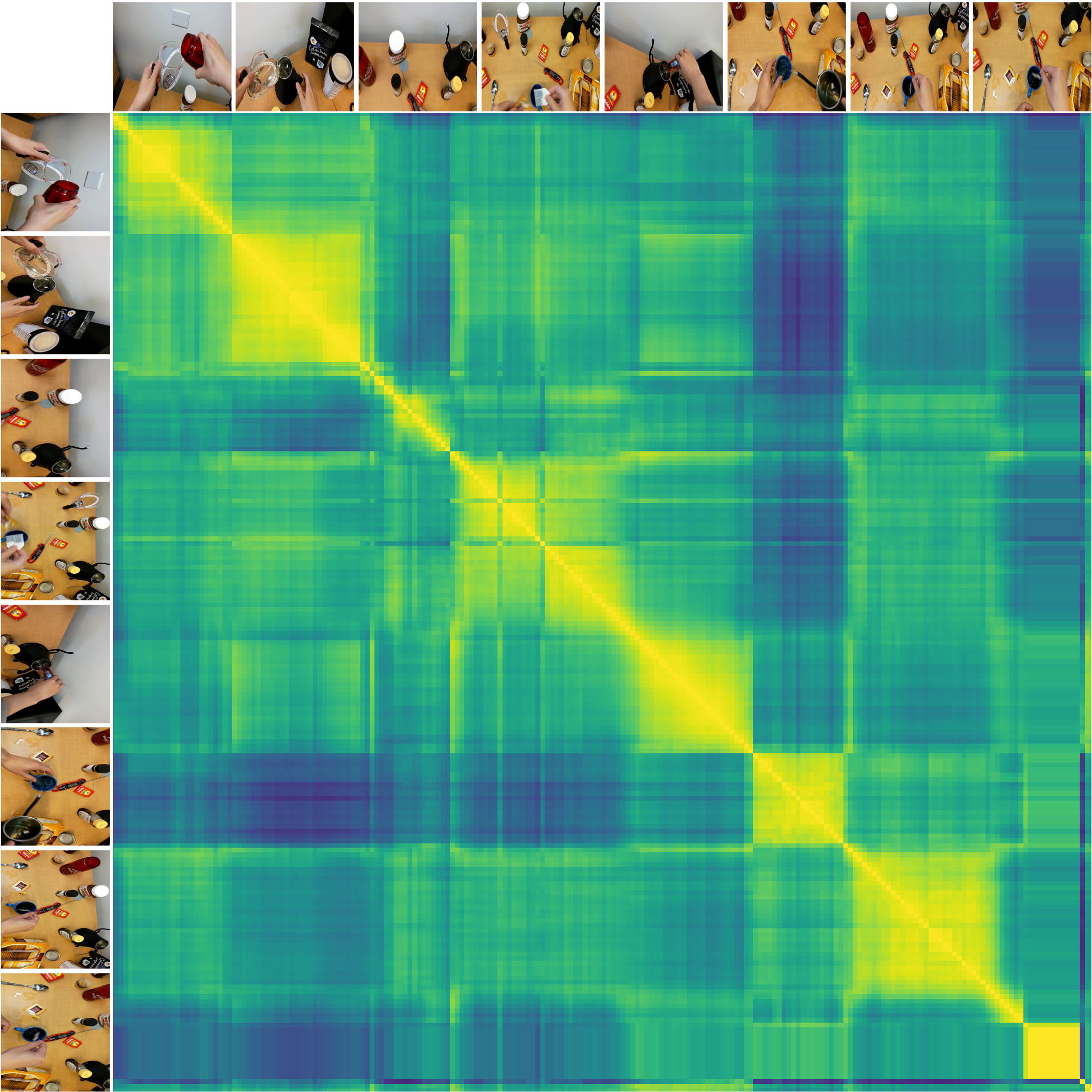}
    \caption{\textbf{Boundary Detection via TSM}}
    \label{fig:tsm_boundary}
\end{figure}

To provide deeper intuition into our training-free temporal segmentation module, Figure \ref{fig:tsm_boundary} visualizes the Temporal Similarity Matrix (TSM) computed over an egocentric video sequence of making tea. Because our Manipulation-Anchored Features (MAFs) rely on the inherent semantic stability of a frozen DINOv2 backbone, continuous procedural actions naturally manifest as distinct, highly correlated block-diagonal structures within the matrix. As illustrated in the center of the figure, frames belonging to the same atomic action (e.g., measuring the water temperature through a thermometer) exhibit high cosine similarity. Conversely, the off-diagonal regions highlight the low similarity across differing procedural steps. As a user transitions between actions—such as shifting from placing a tea bag to pouring hot water—the hand-object state changes, causing a sharp, rapid decrease in local temporal similarity. By isolating these structural transition points, VidParse successfully partitions the continuous video stream into discrete, semantically coherent segments without requiring any learned temporal filters.

\section{Action conditioned background suppression}
In the main text, we describe a visibility prior that assigns a penalty to segments lacking reliable hand-object interaction evidence. While this effectively filters out background noise, certain procedural tasks naturally involve extended periods where hands leave the frame (e.g., waiting for an appliance or baking).

To maintain procedural continuity during these specific steps, we implement a targeted background suppression mechanism. When the decoder predicts a predefined, recipe-specific non-manual action, it temporarily enters a suppressed state. During this phase, the hand-visibility penalty is explicitly zeroed out, preventing the model from spuriously transitioning to the BG class simply due to a lack of hand detections. The transition cost on the beam is extrapolated linearly based on the action's previously accumulated cost rate. This suppression state acts as an action-conditioned inertia. It is automatically lifted as soon as hand-object interactions become highly visible again, allowing standard visual evidence to drive the inference sequence once more. In our reported results, we have applied this setting only to the action - \textit{Microwave for X seconds} in the EgoPer dataset.

\section{Failure Cases and Limitations}
While VidParse achieves robust, training-free parsing, its design introduces a few specific limitations and potential failure modes. Our structured prior relies on a directed task graph induced directly from training sequences. Because our beam search assigns an infinite cost to transitions that violate this graph, the model cannot recover if a user performs a completely novel—yet practically valid—action sequence. In such out-of-distribution executions, the decoder may prune the correct sequence, forcing an incorrect alignment to the known graph. Additionally, the framework's semantic representation is heavily dependent on the Hand-Object Detector (HOD) ~\cite{shan2020understanding}. While our gap rectification and inertia mechanisms handle transient occlusions, prolonged periods of severe hand-object occlusion or purely non-manual actions can cause the visual evidence to degrade. In these extended failure cases, the model may emit spurious predictions or rely too heavily on the last valid feature state.

\end{document}